# The more polypersonal the better -
# a short look on space geometry of fine-tuned layers

Sergej Kudrjashov[0000-0003-1899-4405], Veronika Zykova[0009-0002-5822-7024],
Angelina Stepanova[0009-0002-2627-2904], Jacob Raskind[0009-0006-2439-5088],
and Eduard Klyshinsky[0000-0002-4020-488X]

National Research University "Higher School of Economics"
xenomirant@gmail.com, eklyshinsky@hse.ru

**Abstract.** The interpretation of deep learning models is a rapidly growing field, with particular interest in language models. There are various approaches to this task, including training simpler models to replicate neural network predictions and analyzing the latent space of the model. The latter method allows us to not only identify patterns in the model's decision-making process, but also understand the features of its internal structure. In this paper, we analyze the changes in the internal representation of the BERT model when it is trained with additional grammatical modules and data containing new grammatical features (polypersonality). We find that adding even a single grammatical layer causes the model to separate the new and old grammatical systems within itself, improving the overall performance on perplexity metrics.

## 1  Introduction

A common approach to many NLP tasks nowadays  is to fine-tune pre-trained language models based on the Transformers architecture [1]. As a result, the attention of researchers is highly attracted to their ability to generalize with only relatively small amounts of data from a specific domain. In the case of large language models, we can talk not only about small amounts, but also missing data, since models like ChatGPT are known to be capable of solving problems that were not directly encountered during the learning phase. For example, Wei and co-authors used it to solve the problem of extracting information in a zero-shot setting [2] and found that ChatGPT is sometimes capable of outperforming full-shot models in such tasks.

ChatGPT is sometimes capable of outperforming full-shot models in such tasks.

In addition to the widespread use of Transformers in solving applied problems, another consequence of such interest is a large number of studies dedicated to the problem of understanding a model's decision-making process and its representative internal geometry. For example, Anelli and co-authors [3] investigated the characteristics of the latent space inside the BERT model from the point of knowledge graphs in order to understand the type and form of information contained in the model.

One of the areas in research on the geometry of language models' internal space is the study of their adaptation to new or unusual grammatical situations and categories. For example, Chistyakova and Kazakova, in their work [4], investigated changes in BERT learning process and performance in a scenario with a "broken" version of Russian, where there is no gender agreement between adjectives. They found that although the quality of the "broken" model slightly decreases according to the goal metric, it is still capable of adaptation to new data properties.

In our work, we aim to investigate the change of a model's latent space w.r.t. an unexpected new grammatical feature. We will do this by adding a new feature consistently and logically to the textual data, rather than contrasting a "broken" and "normal" model. Our goal is to leave the model with the opportunity to learn a new system, and to explore the learning outcomes in terms of a target metric and internal geometry of the latent space. By selecting a specific grammatical category and consistently changing it in the data while keeping everything else unchanged, we will be able to fine-tune a selected model (it is also BERT as we are following the stream of aforementioned research) and draw conclusions about its ability to adapt to the new grammatical context.

## 2 Literature Review

### 2.1 Interpretable Machine Learning

The field of interpretable machine learning (IML), also known as explainable artificial intelligence, has a long history. Despite this fact even the definition of these terms remains debatable. For example, Murdoch with co-authors [5] defines IML as the ability to extract "relevant knowledge" from the machine-learning model (where relevant means providing an essential insight appropriate for the particular audience). Another definition can be found in the paper of Doshi-Velez and Kim [6] where they understand interpretability as "the ability to explain or to present in understandable terms to a human". It is easy to see that while the first variant has nothing to do with the ground causes of concrete decisions; the second one operates with the terms of explanations and understanding. However, there are many researches postulating the latter to be a separate concept named "explainability".

Although the variety of definitions makes utilizing such concepts difficult, there is much research dedicated to this topic written throughout the lifetime of machine learning. In the beginning they were mostly connected to the statistical origin of the "classical" machine learning methods such as linear regression or SVM. Then it was the interpretability that provided the reason for using models due to their ability to discover connections undetectable by humans. Although such models are initially interpretable, there have been several studies on ways to keep them so in cases of manipulation with data or model structure. For example, Flanders with co-authors [7] provides a method of interpreting results of linear regression models in case of dependent variable transformation.

IML has been a popular topic since the very beginning of machine learning. However, it has become particular importance since the dramatic increase in development of neural networks and especially large language models. Nowadays models are often treated as black boxes with an unexplainable decision-making logic that contradict the main ideas of IML. For this reason, there is an increasing demand in research devoted to the interpretation and identifying the boundaries of awareness of such models. For example, Clark with co-authors [8] made an attempt to provide attention mechanism interpretation and Hewitt and Manning [9] probed BERT for its syntax understanding.

### 2.2 Latent space geometry

The study of latent space geometry is a broad field of research concerned with finding model internal structures that guide the model learning and "thinking" process. However, these structures can be arbitrarily complex and noisy – especially when the model is not regularized and has less intrinsic biases.

Therefore various techniques that aim to explore these representations' geometry have been proposed. Cai et al [10] explores the isotropy of contextual embeddings layerwise and find that although transformer embeddings tend to clusterize within a cone [11], they keep locally isotropic, and their further regularization and whitening improves their quality even further [12]. Coenen and co-authors [13] find evidence that BERT fine-grained representations of semantics and syntax are encoded in a different manner. While semantics are encoded in a low-dimensional space, syntax tends to be expressed in a tree-structure encoded via associating directions in the latent space with the respective syntactic relations.

There have also been various attempts to study contextual language models' representations using algorithms from topological data analysis. Rathore and co-authors [14] explore the change in embeddings on various layers during fine-tuning and find that "fine-tuning changes the topological structures of embeddings in higher layers more than in lower layers" which is in line with attributing the main interpreting function to them while lower layers serve as a feature encoder and selector. Meirom and Bobrowski [15] encode multilingual sentences in terms of their equivalent translated sentence distance matrices where each matrix encodes distances between embeddings of the respective words, and show that "geometric and topological structures of sentences are preserved to a significant level across languages".

## 3 Experiments

### 3.1 Model Architecture

We experimented with two models. The first model consists of two blocks. The first block is RuBERT pre-trained on the standard version of the training corpus in advance. This block is frozen during training as we do not want RuBERT to acquire information about polypersonality.

The second block is an MLM head with two linear layers (projection matrix and an output embedding layer namely) with a GeLU activation function and a layer norm in between. The output embedding weights are tied to the RuBERT input embeddings, with only the bias term being trained. According to Inan et al. [16], the projection of model representations onto the embedding matrix from the first layers does not cause a decrease and may even improve the model performance in deep autoregressive language models while keeping the number of parameters less by a substantial amount. This helps make the training process more efficient, significantly reducing memory usage.

The second model includes an extra layer placed in between the BERT and the projection matrix. It is an LSTM layer, which is supposed to model and interpret information about polypersonality better than general MLP. It is present only in the second model, as we wanted to see if it is necessary to train a more complex model or training a model with just one linear layer before the output embeddings would be sufficient. It also provides us with a comparison for change in representation geometry between various model layers given the same input.

## 3.2 Data

To conduct our experiments, we created a set of minimal pairs. These pairs consist of an example from a standard language and an example that has been generated by adding a grammatical feature to it.

The first step in our study was to determine which grammatical category to add. To draw confident conclusions about feature localization, this feature has to be marked unambiguously and be absent from the language. Moreover, the addition of this category should not result in creation of new tokens as this requires training a new tokenizer. As a result, we chose adding polypersonal agreement to Russian.

In languages with polypersonal agreement, the verb agrees with more than one of its arguments. To create a modified version of the Russian language that includes polypersonal agreement, we have added an affix to each transitive verb that has a direct object. The affix is present only if the object is expressed as a noun or pronoun.

The affix is chosen based on the person and number of the object. For example, to add polypersonality to the sentence она слышала каждое слово (ona slishala kazhdoje slovo) (tr. she heard every word),we add affix -ет (-jet) as a suffix: она слышалает каждое слово (ona slishalajet kazhdoje slovo).

These affixes vary depending on the conjugation of the verb. Before choosing the affixes, we analyzed the behavior of the BERT-Small tokenizer and found a paradigm that causes minimal changes to the tokenization of verbs, which turned out to be the first conjugation paradigm.

We also experimented with another approach when the same affix is added as the prefix of the verb, for example, она етслышала каждое слово (ona jetslishala kazhdoje slovo). We suppose that as the number of words that start with verb subject affixes is low, it may present rather more unusual setting for the model both in terms of tokenization, sequence length and linguistic preferability (both phonetically and syntactically as inflectional categories are generally encoded suffixally in Russian) while expressing the same linguistic category.

Note that we could have experimented with any other grammatical feature. However, we aimed to seek for changes in resulting BERT latent space geometry without changing its tokenizer but still providing it with a completely new grammatical feature for the selected language. That is why we found polypersonal agreement as a suitable option for our purposes.

The data we used were obtained from a collection of fiction texts from the Internet.

### 3.3 Experiment Design

At first, in order to adapt the model to our data, RuBERT was pretrained for 10 epochs with all the layers unfrozen on MLM task using regular Russian sentences and modified polypersonal Russian sentences mixed in equal amounts. To localize models' knowledge and change in its representations of the selected grammatical feature, we propose an experiment.

We design settings that differ by a parameter and take measurements in each of them respectively: base and polypersonal sentences with another split in the position of polypersonal affix – at the very beginning of the word (prefix position) or at the end (suffix position). It should be noted that suffix and prefix settings are represented by the independent models trained on the respectively preprocessed texts. The equality of training parameters, architecture and the training set besides difference in polypersonal agreement marking lets us compare them both in terms of perplexity and representations.

After training we measured the pseudo-perplexity of the obtained model on a 10.000 sentence sample from a test dataset using PLL-word-l2r scorer [17] where the perplexity of the words tokenized into multiple parts is measured with the right parts masked. As noted in [17], it provides a more accurate estimate of perplexity on multi-token and low-frequency words. This is crucial for us as most of our target words are indeed multi-token. The estimated perplexity of a sentence is an average of its constituent tokens.

We also sample 1000 sentences from the test dataset to estimate a difference in vector representations between settings and outputs of model layers. We suppose that the trained block is capable of better differentiating base and polypersonal sentences in terms of their representations in its latent space. The estimate of layer-setting difference is given by the bottleneck distance [18] between persistent diagrams constructed using Vietoris-Rips filtration [19] for each token vector representation taken after the respective layer. We compare them against each other together with the pretrained model outputs before polypersonal fine-tuning. The final estimate is given by averaging over all the sentences.

The bottleneck distance gives us an estimate on the max norm difference of layer-setting topology that is persistent with respect to small perturbations and noise in the internal latent space [18].

Also, 2000 samples were chosen for layer-wise analysis of polypersonal inflection prediction using Ecco [20]. In this experiment, models' inference was performed on samples with masked polypersonal inflections. Each layer outputs were passed through either:

a) fine-tuned MLM head,
b) combination of LSTM and MLM head.

We analyzed the probability and rank of the correct inflection token for prediction in each setup.

## 4 Results

### 4.1 Perplexity

In terms of perplexity (results are summarized in figure 1 below), there is a general pattern across all variants of the model, including the initial checkpoint: perplexity in texts with polypersonality is higher than in standard Russian. However, for models fine-tuned within the framework of the experiment, the difference between the metrics has been reduced, which, along with a decrease in standard deviation, allows us to speak about a successful, although incomplete, adaptation of the model to a new category.

The results are generally consistent with our assumptions. LSTM is capable of identifying polypersonality and prefix marking tends to be more difficult for models to process.

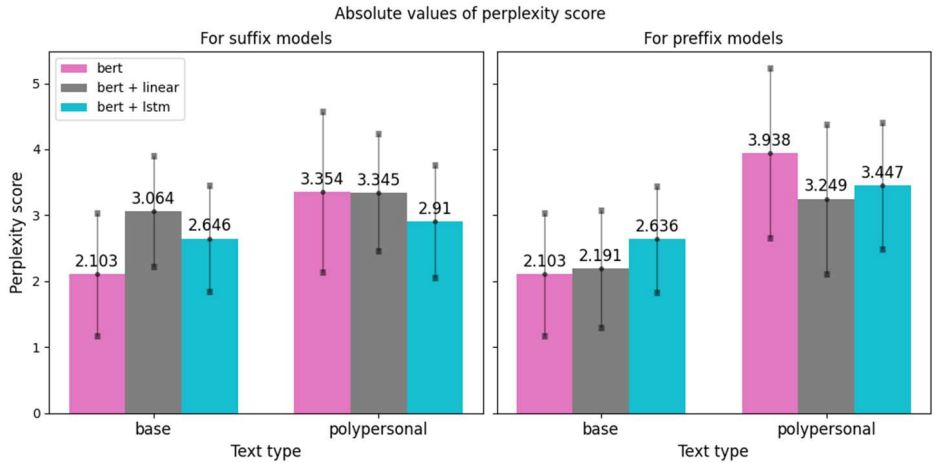

**Fig. 1.** Perplexity score for described models

### 4.2 Probing for MLM on intermediate layers

Experimental results show several interesting patterns. For fine-tuned heads the probability of correct polypersonal inflection prediction starts to progressively grow from the middle layers with an obvious peak at the last for the suffix models. MLP model also has an interesting peak on earlier layers which is absent for LSTM+MLP (Fig, 2). Prefix models are remarkably different with a rapid growth to high confidence on the sixth layer with an overall plateau after. As inflectional prefixes of these forms are absent from Russian we suppose that the system, while being harder for generalization for the whole language, which is shown by PPL evaluation, provides an easier local verbal subsystem to learn on the token level. However, the layer distribution with such explosive growth is notable.

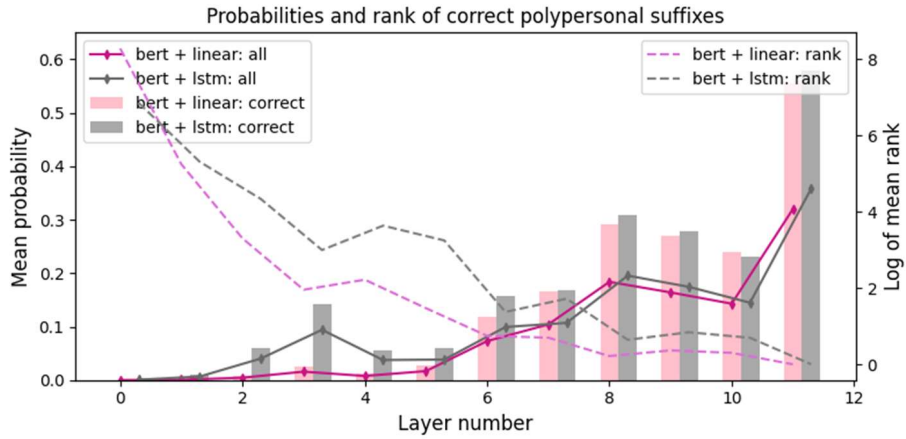

**Fig. 2.** Probability of correct token and its rank probed on BERT layers

### 4.3 Latent Space Geometry

The results of the study of models' internal space geometry are summarized in two figures (for models with a linear layer and an LSTM block, respectively). Here, the types of models are arranged vertically and horizontally: pref/suff - shows whether the prefix or suffix versions of polypersonality were checked, base/poly - shows whether the input text was polypersonal, bert/gram - shows whether the outputs of the BERT or grammar module are used for calculations. It should be noted that the outputs of BERT are common to all variants of models, which is why it is considered once for specific characteristics of texts.

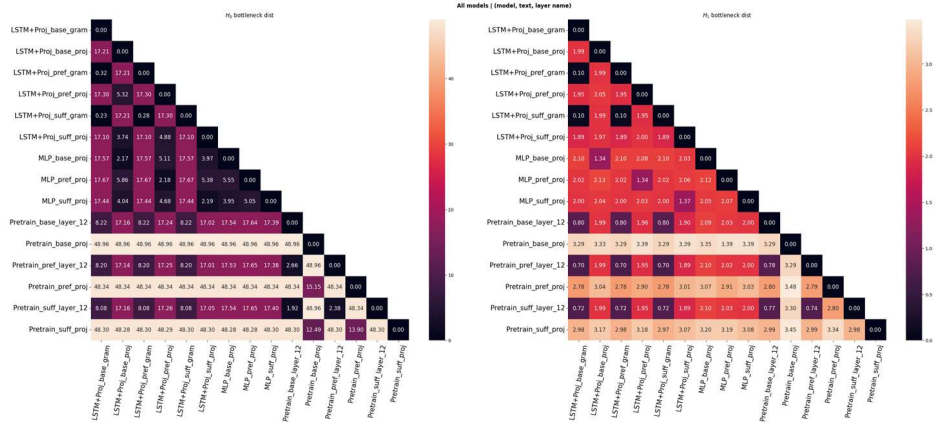

**Fig. 3.** Comparison of model and text bottleneck distances, $H_0$ and $H_1$ groups

The most prominent feature that strikes the first is the difference between pre-trained BERT layer outputs and fine-tuned ones. However, BERT embeddings are quite similar within themselves when the input text is polypersonal. This suggests that BERT does distinguish between different strategies of polypersonality, as indicated by the results of the perplexity assessment and the non-zero values in its comparisons. However, it does not seem to do so to a sufficient extent and they tend to be rather similar as the BERT never actually encountered polypersonal sentences during pretraining. The grammar module, on the other hand, seems to cope with this by disentangling texts with different grammar in a rather diverse way. This allows us to conclude that the model, while leaving the geometry describing standard russian-language texts unchanged, significantly shifted texts with polypersonality relatively to them.

It should be also noted that despite the aforementioned differences between suffix and prefix models, they are not that different in terms of topological structure with a slight higher similarity between suffix and base models against prefix ones. This is consistent across models. However, the clear identification of the sources is difficult. We suggest that it may be explained both by tokenization (hence positional encoding effects) and linguistic improbability for the pretrained model. The extent to which each factor contributes is prior for further work.

Another striking pattern comes from a comparison between MLP and LSTM layer taken. Although LSTM is characterized with the same tendency to disjoin polypersonal representations from base ones, the distances between them are extremely smaller. We suggest that it may happen due to the need of the recurrent model to preserve some consistent representations without any of them exploding due to high variance. However, this idea needs further refinement and research in the future [21].

As a result, we can conclude that the fine-tuning of a certain layer before the prediction head lets BERT learn how to differentiate strategies of polypersonality while leaving the general understanding of the language mostly unchanged.

# 5    Conclusion

We studied the geometry of an added fine-tuned layer of a BERT model on a mix of two languages: general and augmented Russian with the added polypersonal affix, thus differing by one linguistic parameter. As a result we found several properties characterizing the process of BERT adaptation to the new context of the input data.

Firstly, BERT with a fine-tuned layer is capable of processing both verbs with polypersonal affixes and the regular Russian language. Although perplexity on polypersonal text is higher for both fine-tuned and initial models, the difference is greater on the model without the new layer.

Secondly, we can see that although BERT has some internal capabilities to differentiate between languages varying in one linguistic parameter, fine-tuning layers creates disentangled and linearly separable representations by changing the representation topology, thus providing the model with more robust features for further processing. Furthermore, while the results with both prefixes and suffixes are relatively similar (the prefix ones differ slightly more), we can see an noticable pattern characteristic for LSTM layers. The later demonstrates significantlymore clustered geometry

Thirdly, probing  the layers with a classification head attached to them, one by one, the further you go, the better the performance you get. The average probability of a correct token increases, and the rank of the correct answer tends towards 1. This is in line with what we see in VisBert [22], where the authors say that already the fifth layer is responsible for establishing connections between words. This looks similar to the increase we can observe  in our data. Since polypersonality is a feature that provides syntactic connections, this similarity seems to be expected.

Finally, our approach allows enhancing the structure of the BERT model by adding new blocks independently trained on a new grammatical feature processing. Note that our approach could not be scaled to other models such as T5 or GPT since perplexity could not be directly applied to inner layers of mentioned models.

# References

1. Vaswani, A., Shazeer, N., Parmar, N., Uszkoreit, J., Jones, L., Gomez, A. N., ... & Polosukhin, I. (2017). Attention is all you need. Advances in neural information processing systems, 30.
2. Wei, X., Cui, X., Cheng, N., Wang, X., Zhang, X., Huang, S., ... & Han, W. (2023). Zero-shot information extraction via chatting with chatgpt. arXiv preprint arXiv:2302.10205.
3. Anelli, V. W., Biancofiore, G. M., De Bellis, A., Di Noia, T., & Di Sciascio, E. (2022, October). Interpretability of BERT latent space through knowledge graphs. In Proceedings of the 31st ACM International Conference on Information & Knowledge Management (pp. 3806-3810).
4. Chistyakova, K. E., & Kazakova, T. B. (2023). Grammar In Language Models: Bert Study (No. WP BRP 115/LNG/2023). National Research University Higher School of Economics.
5. Murdoch, W. J., Singh, C., Kumbier, K., Abbasi-Asl, R., & Yu, B. (2019). Definitions, methods, and applications in interpretable machine learning. Proceedings of the National Academy of Sciences, 116(44), 22071-22080.

6. Doshi-Velez, F., & Kim, B. (2017). Towards a rigorous science of interpretable machine learning. arXiv preprint arXiv:1702.08608.

7. Flanders, W. D., DerSimonian, R., & Freedman, D. S. (1992). Interpretation of linear regression models that include transformations or interaction terms. Annals of Epidemiology, 2(5), 735-744.

8. Clark, K., Khandelwal, U., Levy, O., & Manning, C. D. (2019). What does bert look at? An analysis of bert's attention. arXiv preprint arXiv:1906.04341.

9. Hewitt, J., & Manning, C. D. (2019, June). A structural probe for finding syntax in word representations. In Proceedings of the 2019 Conference of the North American Chapter of the Association for Computational Linguistics: Human Language Technologies, Volume 1 (Long and Short Papers) (pp. 4129-4138).

10. Xingyu Cai, Jiaji Huang, Yuchen Bian, & Kenneth Church (2021). Isotropy in the Contextual Embedding Space: Clusters and Manifolds. In International Conference on Learning Representations.

11. Jun Gao, Di He, Xu Tan, Tao Qin, Liwei Wang, & Tieyan Liu (2019). Representation Degeneration Problem in Training Natural Language Generation Models. In International Conference on Learning Representations.

12. Junjie Huang, Duyu Tang, Wanjun Zhong, Shuai Lu, Linjun Shou, Ming Gong, Daxin Jiang, & Nan Duan (2021). WhiteningBERT: An Easy Unsupervised Sentence Embedding Approach. arXiv preprint arXiv: 2104.01767.

13. Andy Coenen, Emily Reif, Ann Yuan, Been Kim, Adam Pearce, Fernanda Viégas, & Martin Wattenberg (2019). Visualizing and Measuring the Geometry of BERT. arXiv preprint arXiv: 1906.02715.

14. Archit Rathore, Yichu Zhou, Vivek Srikumar, & Bei Wang (2023). TopoBERT: Exploring the topology of fine-tuned word representations. Inf. Vis., 22(3), 186–208.

15. Haim Meirom, S., & Bobrowski, O. (2022). Unsupervised Geometric and Topological Approaches for Cross-Lingual Sentence Representation and Comparison. In Proceedings of the 7th Workshop on Representation Learning for NLP (pp. 173–183). Association for Computational Linguistics.

16. Hakan Inan, Khashayar Khosravi, & R. Socher (2016). Tying Word Vectors and Word Classifiers: A Loss Framework for Language Modeling. International Conference on Learning Representations.

17. Kauf, C., & Ivanova, A. (2023). A Better Way to Do Masked Language Model Scoring. In Proceedings of the 61st Annual Meeting of the Association for Computational Linguistics (Volume 2: Short Papers) (pp. 925–935). Association for Computational Linguistics.

18. David Cohen-Steiner, Herbert Edelsbrunner, & John Harer (2005). Stability of Persistence Diagrams. Discrete & Computational Geometry, 37, 103-120.

19. Aken, B. V., Winter, B., Löser, A., & Gers, F. A. (2020, April). Visbert: Hidden-state visualizations for transformers. In Companion Proceedings of the Web Conference 2020 (pp. 207-211).